\documentclass[11pt]{article}
\usepackage{acl2015}
\usepackage{times}
\usepackage{url}
\usepackage{latexsym}
\usepackage{multirow}
\usepackage{graphicx,epstopdf}
\usepackage{amsmath,amssymb,bm}
\usepackage{breqn}
\usepackage{ulem}
\usepackage{subfigure}
\usepackage{bm}

\title{Topic2Vec: Learning Distributed Representations of Topics}

\author{Li-Qiang Niu and Xin-Yu Dai \\
 National Key Laboratory for Novel Software Technology, Nanjing University, Nanjing, China \\
  {\tt \{niulq, daixinyu\}@nlp.nju.edu.cn} \\}

\date{}

\begin{document}
\maketitle
\begin{abstract}

Latent Dirichlet Allocation (LDA) mining thematic structure of documents plays an important role in nature language processing and machine learning areas. However, the probability distribution from LDA only describes the statistical relationship of occurrences in the corpus and usually in practice, probability is not the best choice for feature representations. Recently, embedding methods have been proposed to represent words and documents by learning essential concepts and representations, such as Word2Vec and Doc2Vec. The embedded representations have shown more effectiveness than LDA-style representations in many tasks. In this paper, we propose the Topic2Vec approach which can learn topic representations in the same semantic vector space with words, as an alternative to probability. The experimental results show that Topic2Vec achieves interesting and meaningful results.

\end{abstract}

\section{Introduction}

Modeling text (words, topics and documents) is a key problem in nature language processing (NLP) and information retrieval (IR). The goal is to find short and essential descriptions which enable efficient processing of large systems and benefit basic tasks such as classification, clustering, summarization and estimation of similarity or relevance.

During the past decades, various models and solutions are proposed, such as Bag-of-Words (BOW) ~\cite{Harris:1954}, {\em TF-IDF} ~\cite{Salton:1983}, Latent Semantic Analysis (LSA) ~\cite{Landauer:1998} and Probabilistic Latent Semantic Analysis (PLSA) ~\cite{Hofmann:1999}. But the best-known model is Latent Dirichlet Allocation (LDA) ~\cite{Blei:2003} which describes the hierarchical relationships between words, topics and documents. In LDA, documents are represented as probability distributions over latent topics where each topic is characterized by a distribution over words. However, the probability distribution generated from LDA prefers to describe the statistical relationship of occurrences rather than real semantic information embedded in words, topics and documents. Also LDA will assign high probabilities to high frequency words and those words with low probabilities are hard to be chosen as representatives of topics. But in practice, low probability words sometimes distinguish topics better.  For example, LDA will assign higher probability and choose ``{\it food}" as representative other than ``{\it cheeseburger}", ``{\it drug}" other than ``{\it aricept}" and ``{\it technology}" other than ``{\it smartphone}".

Recently, distributed representations with neural probabilistic language models (NPLMs) ~\cite{Bengio:2003} were proposed to represent words and documents as low-dimensional vectors in one semantic space, and achieved significant results in many NLP and ML tasks ~\cite{CW:2008,MH:2009,Mikolov:2013a,MK:2013,Huang:2012,Le:2014}. In particular, Word2Vec proposed by ~\newcite{Mikolov:2013a} could automatically learn concepts and semantic-syntactic relationships between words like vec(``{\it Berlin}") - vec(``{\it Germany}") = vec(``{\it Paris}") - vec(``{\it France}"). Doc2Vec (Para2Vec) proposed by ~\newcite{Le:2014} achieves state-of-the-art performance on sentiment analysis. Naturally, in this paper, we want to answer the question that, what will happen if we embed topics in the semantic vector space?

Following the ideas of previously proposed models for words and documents, we propose the model Topic2Vec as shown in Fig. \ref{fig:our_model}. Based on the Word2Vec, we incorporates topics into the NPLM framework for learning distributed representations of topics in the same semantic space with words. Furthermore, words and topics naturally can estimate similarity and relevance with each other such as using cosine function rather than using probability.

In the experiments, we evaluate two different topic representations including embedding of Topic2Vec and probability of LDA in two aspects: listed examples and t-SNE 2D embedding of nearest words for each topic. The experimental results show that our Topic2Vec achieves distinctive and meaningful results compared to LDA.

\section{Related Models}
\label{sec:bg}

\subsection{Latent Dirichlet Allocation}
\label{subsec:bg_lda}

Latent Dirichlet allocation (LDA) \cite{Blei:2003} is a probabilistic generative model that assumes each document is a mixture of latent topics, where each topic is a probability distribution over all words in vocabulary. Briefly, LDA generates a sequence of words as follows:
\begin{itemize}
\item For each of the {\em N} word $w_{n}$ in document $d$:
\begin{itemize}
\item Sample a topic $z_{n}$ $\sim$ Multinomial($\theta_{d}$)
\item Sample a word $w_{n}$ $\sim$ Multinomial($\phi_{z_{n}}$).

\end{itemize}
\end{itemize}

By Gibbs Sampling \footnote{http://gibbslda.sourceforge.net/} estimation, we obtain document-topic probability matrix $\Theta$ and topic-word probability matrix $\Phi$. For a new document of arbitrary length, we can infer its involved latent topics and meanwhile we will assign a topic label for each word in the document.

\subsection{Word2Vec}
\label{subsec:bg_word2vec}

Inspired by Neural Probabilistic Language Model (NPLM) ~\cite{Bengio:2003}, ~\newcite{Mikolov:2013a} proposed Word2Vec including CBOW and Skip-gram for computing continuous vector representations of words from large data sets.

When training, given a word sequence $D=\left \{w_{1},...,w_{M}  \right \}$, the learning objective functions are defined to maximize the following log-likelihoods, based on CBOW and Skip-gram, respectively.
\begin{dgroup}
\begin{dmath}
\mathcal{L}_{CBOW}(D)=\frac{1}{M}\sum_{i=1}^{M}\log p(w_{i}|w_{cxt}),
\label{eq:wv-cbow}    
\end{dmath}
\begin{dmath}
\mathcal{L}_{Skip-gram}(D)=\frac{1}{M}\sum_{i=1}^{M}\sum_{-k\leq c\leq k,c\neq 0}\log p(w_{i+c}|w_{i}).
\label{eq:wv-skip}    
\end{dmath}
\end{dgroup}
Here, in Equation (\ref{eq:wv-cbow}), $w_{cxt}$ indicates the context of the current word $w_{i}$. In Equation (\ref{eq:wv-skip}), $k$ is the window size of context. For any variables $w_{j}$ and $w_{i}$, the conditional probability $p(w_{j}|w_{i})$ is calculated using softmax function as follows,
\begin{equation}
\label{softmax}
p(w_{j}|w_{i})=\frac{\exp(\mathbf{w_{j}} \cdot \mathbf{w_{i}})}{\sum_{w\in W}\exp(\mathbf{w} \cdot \mathbf{w_{i}})},
\end{equation}
where $\mathbf{w}$, $\mathbf{w_{i}}$ and $\mathbf{w_{j}}$ are respectively the word representations of word $w$, $w_{i}$ and $w_{j}$, $W$ is the word vocabulary.

\begin{figure}[t]
\centering
\includegraphics[height=3.8cm,width=3.2in]{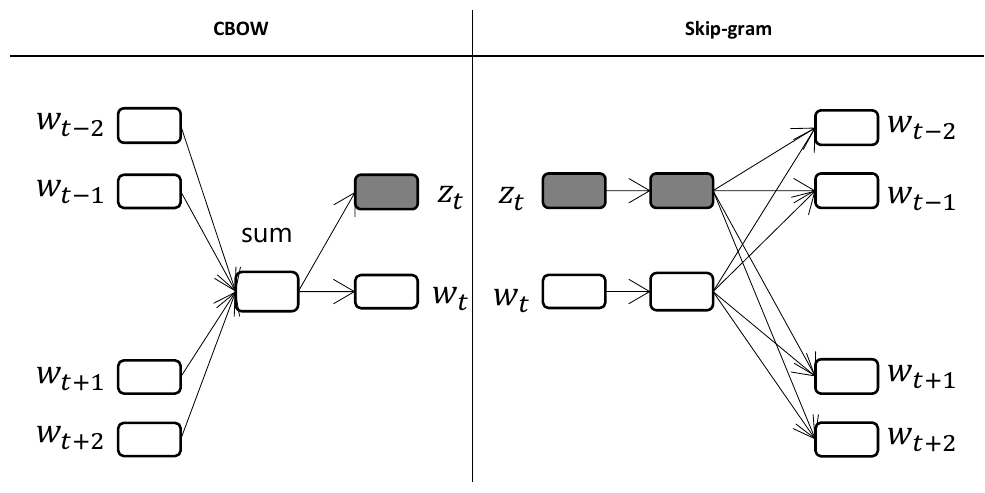}
\caption{Learning architectures of Topic2Vec, where $(w_{t-2},w_{t-1},w_{t+1},w_{t+2})$ are context words and $w_{t}$ is the current word paired with a topic $z_{t}$.}
\label{fig:our_model}
\end{figure}

\begin{figure*}[t]
\centering
\includegraphics[height=12.8cm,width=6.2in]{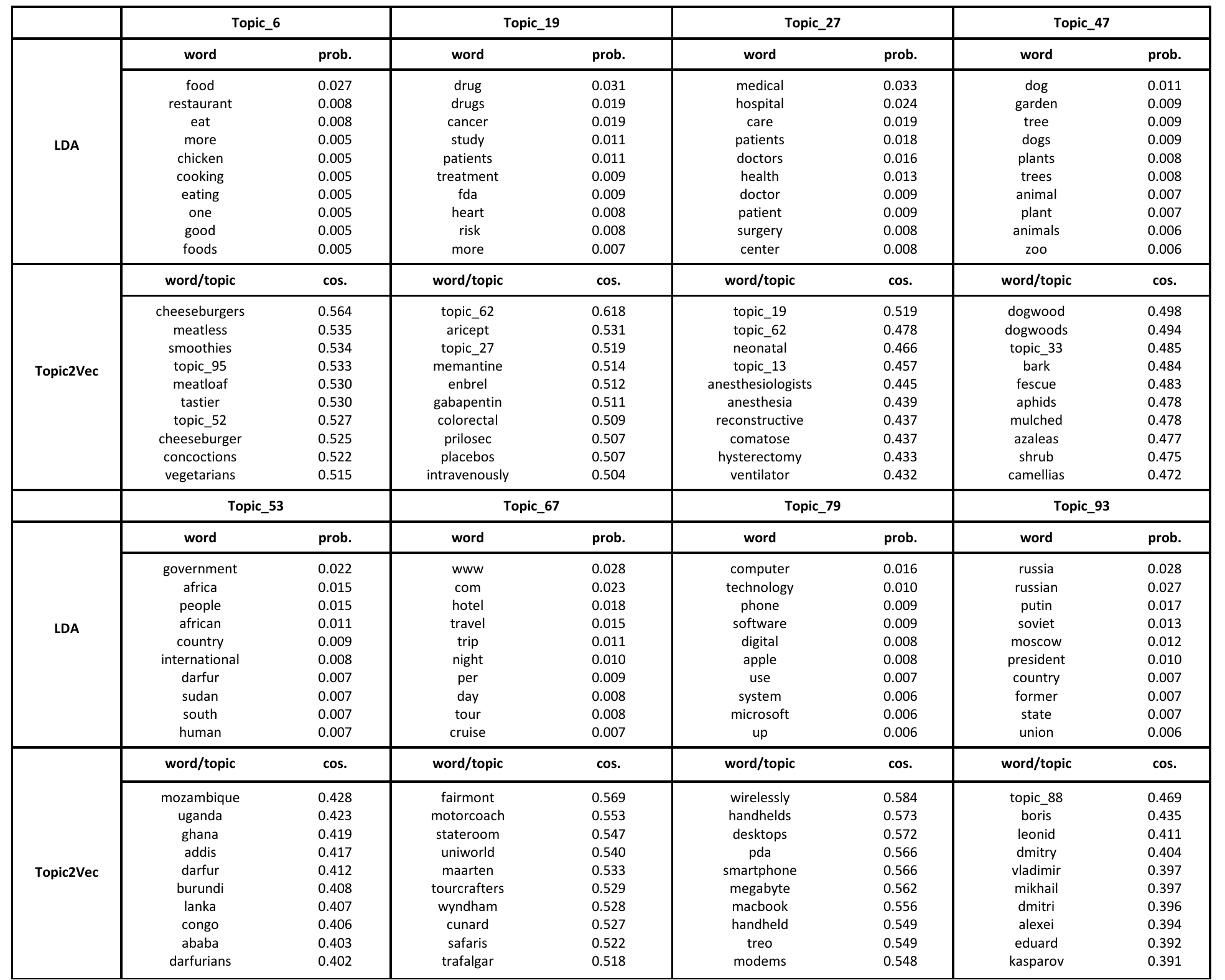}
\caption{Nearest words and topics for each topic. Words are listed with conditional probabilities in LDA while words and topics are listed with calculated cosine similarity in Topic2Vec.}
\label{fig:exp_nearest}
\end{figure*}

\section{Topic2Vec}
\label{sec:topic2vec}

Inspired by word2vec, we incorporate topics and words into the NPLM. We propose Topic2Vec as shown in Fig. \ref{fig:our_model} for learning distributed topic representations together with word representations. Topic2Vec is also separated in CBOW and Skip-gram situations. For instance, given a word sequence $(w_{t-2}, w_{t-1}, w_{t}, w_{t+1}, w_{t+2})$, in which $w_{t}$ is the current word assigned with topic $z_{t}$ by LDA. The CBOW predicts the word $w_{t}$ and topic $z_{t}$ based on the surrounding words $(w_{t-2},w_{t-1},w_{t+1},w_{t+2})$, while the Skip-gram predicts surrounding words $(w_{t-2},w_{t-1},w_{t+1},w_{t+2})$ given current $w_{t}$ and $z_{t}$.

When training, given a word-topic sequence of a document $D=\left \{w_{1}:z_{1},...,w_{M}:z_{M}  \right \}$, where $z_{i}$ is the word $w_{i}$'s topic inferred from LDA, the learning objective functions can be defined to maximize the following log-likelihoods, based on CBOW and Skip-gram, respectively.
\begin{dgroup}
\begin{dmath}
\mathcal{L}_{CBOW}(D)=\frac{1}{M}\sum_{i=1}^{M}(\log p(w_{i}|w_{cxt})+\log p(z_{i}|w_{cxt})),
\label{eq:our_cbow}    
\end{dmath}
\begin{dmath}
\mathcal{L}_{Skip-gram}(D)=\frac{1}{M}\sum_{i=1}^{M}\sum_{-k\leq c\leq k,c\neq 0}(\log p(w_{i+c}|w_{i})+\log p(w_{i+c}|z_{i})).
\label{eq:our_skip}
\end{dmath}
\end{dgroup}

Topic2Vec aims at learning topic representations along with word representations. Considering the simplicity and efficient solution, we just follow the optimization scheme that used in Word2Vec ~\cite{Mikolov:2013a}. To approximately maximize the probability of the softmax, we use Negative Sampling without Hierarchical Softmax ~\cite{Mikolov:2013b}. Stochastic gradient descent (SGD) and back-propagation algorithm are used to optimize our model. By the way, complexity of our Topic2Vec is linear with size of dataset, same with Word2Vec.

\begin{figure*}[t]
\subfigure{
\begin{minipage}[b]{3.2in}
\includegraphics[height=4.8cm,width=3.1in]{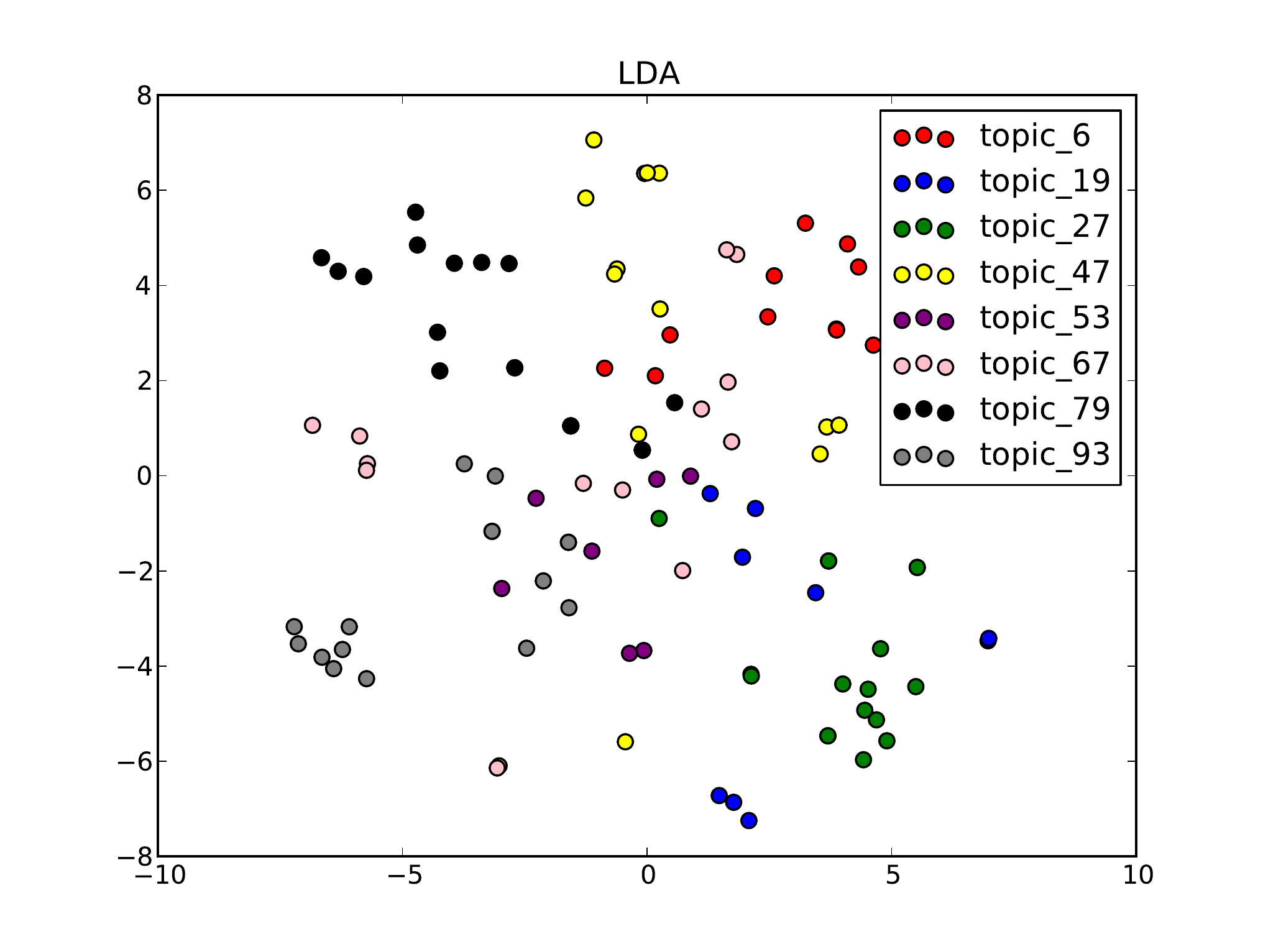}
\label{fig:exp_lda}
\end{minipage}
}
\subfigure{
\begin{minipage}[b]{3.2in}
\includegraphics[height=4.8cm,width=3.1in]{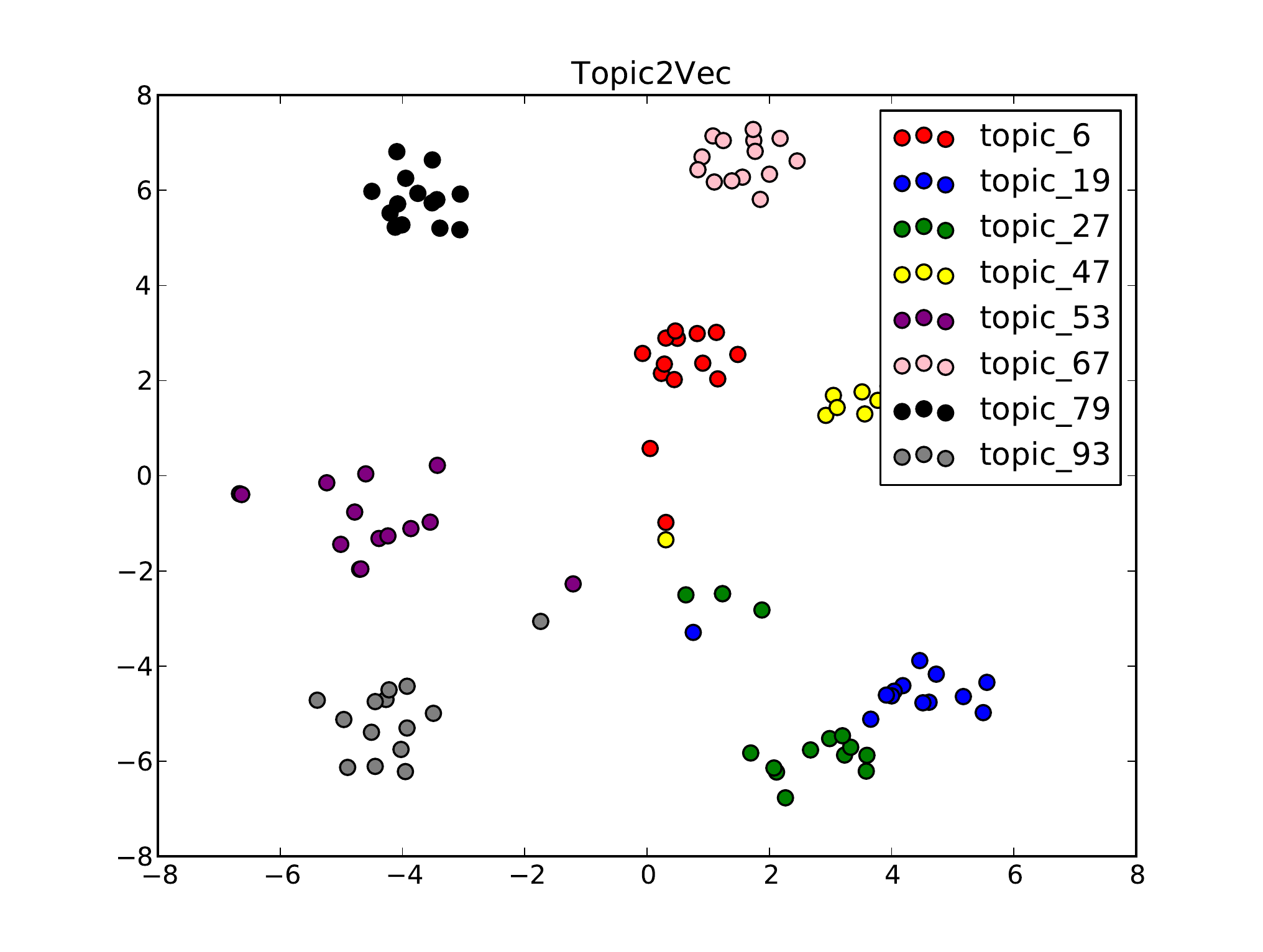}
\label{fig:exp_atr}
\end{minipage}
}
\caption{t-SNE 2D embedding of the nearest word representation for each topic in LDA (left) and Topic2Vec (right).}
\label{fig:exp_embedding}
\end{figure*}

\section{Experiments}
\label{sec:experiments}

\subsection{Dataset}
\label{subsec:ds}
We use the English Gigaword Fifth Edition\footnote{https://catalog.ldc.upenn.edu/LDC2011T07} as our training data for learning fundamental word and topic representations. We randomly extract part of documents and construct our training set described as follows: we chose 100,000 documents, where each consists of more than 1,000 characters from subfolder ltw\_eng (Los Angeles Times) containing 411,032 documents. Besides, we eliminate those words that occur less than 5 times and the stop words. In the end, training set contains about 42 million words and the vocabulary size is 102,644. 

\subsection{Evaluation Methods}
\label{subsec:eva}
In experiments, we run Topic2Vec in Skip-gram and learn topic representations together with word representations. And then we evaluate topic representations via comparing Topic2Vec with LDA in two aspects: (1) we select most related topics or words conditioned on selected topics and (2) we embed these related words or topics in 2D space using t-SNE ~\cite{Maaten:2008}. During the process, we cluster words into topics as follows:

\begin{itemize}
\item LDA: each topic is a probability distribution over words. We select the top $N=10$ words with highest conditional probability.
\item Topic2Vec: topics and words are equally represented as the low-dimensional vectors, we can immediately calculate the cosine similarity between words and topics. For each topic, we select higher similarity words.

\end{itemize}

\subsection{Analysis of Results}
\label{subsec:ana}

Fig. \ref{fig:exp_nearest} shows top 10 nearest words from LDA and Topic2Vec for eight typically selected topics, respectively. We now give more detailed analysis to understand the difference between them. As shown in Fig. \ref{fig:exp_nearest}, in Topic\_19, LDA returns the words like ``{\it drug}", ``{\it drugs}", ``{\it cancer}" and ``{\it patients}", while Topic2Vec returns ``{\it aricept}", ``{\it memantine}", ``{\it enbrel}" and ``{\it gabapentin}". In Topic\_27, LDA returns the words of ``{\it medical}", ``{\it hospital}", ``{\it care}", ``{\it patients}" and ``{\it doctors}", while Topic2Vec returns ``{\it neonatal}", ``{\it anesthesiologists}", ``{\it anesthesia}" and ``{\it comatose}". We only know that Topic\_19 and Topic\_27 share the same topic about ``{\it patients}" or ``{\it medical}", but we can't get their further difference from the results of LDA. But from the result of Topic2Vec, we can easily discover that Topic\_19 focuses on a more specific topic about drugs (``{\it aricept}", ``{\it memantine}", ``{\it enbrel}" and ``{\it gabapentin}"), while Topic\_27 focuses on another specific topic about treatment (``{\it anesthesiologists}", ``{\it anesthesia}" and ``{\it comatose}"), they are absolutely different. Obviously, Topic2Vec presents more distinguished results between two similar topics.

Fig. \ref{fig:exp_embedding} shows the 2D embedding of the corresponding related words for each topic by using t-SNE. Obviously, Topic2Vec produces a better grouping and separation of the words in different topics. In contrast, LDA does not produce a well separated embedding, and words in different topics tend to mix together.

In summary, for each topic, words selected by Topic2Vec are more typical and representative compared to those returned by LDA. Eventually, Topic2Vec can better distinguish different topics.

\section{Conclusions and Future Work}
\label{sec:con}

In this paper, via integrating NPLM, Word2Vec and LDA, we propose the Topic2Vec which successfully embeds latent topics in the same semantic vector space with words. In principle, our purpose clearly aims at learning new fashion topic representation by Topic2Vec. From the observation of experiments, Topic2Vec presents more distinguished results than LDA and we have the conclusion that Topic2Vec can model topics better.

But now, we just qualitatively evaluate the performance of Topic2Vec and LDA, we will quantitatively do more detailed analysis about their difference in the future. Besides, we have to run LDA firstly to assign a topic for each word in the corpus before Topic2Vec. We also will explore new independent topic models which can mine thematic structure of documents as LDA and learn inherent representations and model topics better as Topic2Vec, simultaneously.

\section*{Acknowledgments}

%


\end{document}